# A Web Scale Entity Extraction System

**Xuanting Cai**
Facebook Inc
caixuanting@fb.com

**Quanbin Ma**
Facebook Inc
quanbinma@fb.com

**Pan Li**
Facebook Inc
pli@fb.com

**Jianyu Liu**
Facebook Inc
jianyuliu@fb.com

**Qi Zeng**
Facebook Inc
qizeng@fb.com

**Zhengkan Yang**
Facebook Inc
zhengkan@fb.com

**Pushkar Tripathi**
Facebook Inc
pushkart@fb.com

## Abstract

Understanding the semantic meaning of content on the web through the lens of entities and concepts has many practical advantages. However, when building large-scale entity extraction systems, practitioners are facing unique challenges involving finding the best ways to leverage the scale and variety of data available on internet platforms. We present learnings from our efforts in building an entity extraction system for multiple document types at large scale using multi-modal Transformers. We empirically demonstrate the effectiveness of multi-lingual, multi-task and cross-document type learning. We also discuss the label collection schemes that help to minimize the amount of noise in the collected data.

## 1 Introduction

Content understanding finds myriad applications in large scale recommendation system. One example is ranking content with sparse data (Davidson et al., 2010; Amatriain and Basilic, 2012). In such scenarios, content signals can offer better generalization to overcome cold-start problems (Lam et al., 2008; Timmaraju et al., 2020). Another example is explaining the working theory of the recommendation system to users and regulators (Chen et al., 2019). In such scenarios, content signals can offer human understandable features.

This paper presents an overview of the entity extraction platform we build for our recommendation system. Along the way, we overcome several unique challenges: *Multiple Languages* - since our business operates world wide and supports languages from various countries, it is imperative to build a multi-lingual system; *Multiple Entity Types* - we want to extract multiple types of entities including named entities like people and places, as well as commercial entities like products and brands; *Multiple Document Types* - our system should work across multiple structured document types such as web pages, ads and user generated content; *Scale* - owing to our scale, we need a system that is responsive and resource efficient to process billions of documents per day.

In the subsequent sections, we will review the methodology to collect data and the ideas behind the models. Then we will discuss techniques to deploy these models efficiently.

## 2 Notation and Setup

An **entity** is a human interpretable concept that is grounded in a real world notion. A **mention** is a word or a phrase in the text that refers to an entity. For example, both "Joe Biden" and "Biden" are mentions that refer to the same entity that represents the 46th president of the United States. **Entity extraction** is the task of extracting mentions from a given text and linking them to entities. Each instance of this problem consists of a **structured** document with text attributes like title and description, as well as categorical features and metadata, from which we wish to extract multiple entities. We categorize the entity extraction tasks into **closed-world** task and **open-world** task. The former is applicable when we have a fixed predefined universe of entities, say, topics from Wikipedia; while the latter is needed when such a list is not available e.g. products.

## 3 Open World Extraction

In this section, we discuss the data labeling and the model architecture for open world entity extraction.

### 3.1 Data Labeling

Collecting data for the open world entity extraction presents unique challenges since it entails collecting free-form inputs from raters. We design a widget to let raters highlight spans of text, generating a set of positive mentions per example. Each example is rated by three raters, and we evaluate different ways to combine mentions from different raters into ground truth: **And** - selecting the tokens highlighted by all raters; **Or** - selecting the tokens highlighted by any rater; **Majority** - selecting the tokens highlighted by the majority of raters. We compare the quality of these methods with golden sets produced by in-house experts as shown in Table 1. The **Majority** method provides the best trade-off between precision and recall.

| Method | Exact Match F1 |
|---|---|
| **And** | 0.775 |
| **Or** | 0.706 |
| **Majority** | 0.794 |

Table 1: Exact Match F1 is the F1 score of the aggregated rater labels of extract match compared with the expert labels.

In order to audit and enhance the quality of the labeled data, we prepare detailed instructions on navigating the user interface, task-specific reasoning process, sample tasks elucidating the rules, and explanations for handling corner cases. Additionally, we routinely inject known examples to calibrate external raters against our experts. We periodically remove and retrain raters whose outputs digress significantly from the experts. Furthermore, we also track their consistency with consensus labels to detect outliers. Finally, we also perform some rule-based sanitization to rectify common errors. For example, we find that raters often fail to select all the occurrences of a same piece of text. Thus, we broadcast selected mentions back to the entire input to capture all occurrences.

### 3.2 Modeling

We divide the open world entity extraction task into the extraction stage and the clustering stage.

#### 3.2.1 Extraction Stage

In the extraction stage, We try to find all mentions in a text using a sequence to sequence model. As depicted in Figure 1a, our extraction model is based on a pre-trained cross-lingual language model (Lample and Conneau, 2019). For computation efficiency, we choose a multiple layer perception on top of XLM instead of conditional random field layer (Lafferty et al., 2001). We find that the simple multiple layer perception with take-continuous-positive-blocks decoding in the sequence works good enough to provide high quality mentions.

#### 3.2.2 Semi-supervised Clustering Stage

In the clustering stage, we try to collapse all mentions referring to the same concept to a canonical entity. Intuitively, one can run k-means algorithm on embeddings coming from the extraction stage. However we found that the performance of this approach not acceptable for two reasons: The k-means is based on a uniform distribution assumption which the embeddings do not follow; Embeddings taken from extraction model fail to align with the human interpretation for two mentions being the same concept.

We solve the problem with a semi-supervised graph based approach, where we build a dedicated model as illustrated in Figure 1b to predict links between mentions if they represent the same underlying entity. This model is trained on a dataset specialized in mention concept similarity that we collect separately. We adopt the Siamese neural network architecture in order to scale for processing all pairs between hundreds of millions of documents during graph construction. Then we run Louvain community detection algorithm (Blondel et al., 2008) on the resulting graph to collapse close mentions into an entity. We find that this could significantly improve the quality of the clusters.

## 4 Closed World Extraction

In this section, we discuss the data labeling and the model architecture for closed world entity extraction.

### 4.1 Data Labeling

In an ideal world, we would want our raters to select the mentions freely from input text and attach the corresponding Wikipedia entity to it. But that makes it hard for raters to reach any consensus, and impossible for us to perform any quality control. Instead, we make the task a multiple-choice, where we extract beforehand a list of possible mentions, alongside with their potential Wikipedia link candidates, with the help of a pre-defined dictionary. Now the rater only need to choose all the positive mentions, and their corresponding Wikipedia entity, both from a given list.

Similar to open world, we perform quality analysis on different consensus methods. Here we treat wiki entities selected by 2 out of 5 raters to be our community ground truth. This method, compared against the oracle labels provided by in-house experts, can achieve 80% chance of having all ex-tracted entities being correct, and 70% chance of having all correct entities being extracted. Both number would further increase by 14% if we tolerate one single error. As reference, the F1 score of an individual average rater on this task is 0.68.

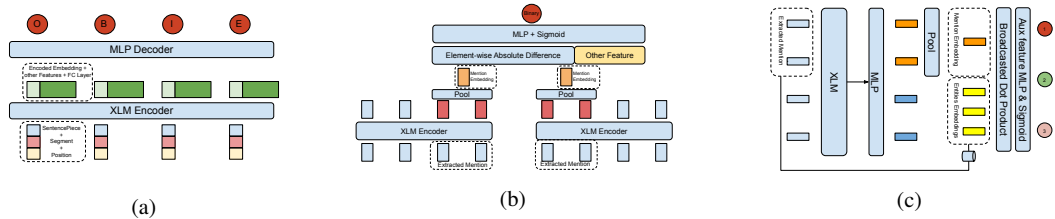

Figure 1: (a) The open world extraction model. (b) The open world link prediction model. (c) The closed-world linking model zooming at one candidate mention. Corresponding entity embeddings are generated offline, and fetched during extraction stage.

### 4.2 Modeling

Similar to the open-world model, we break up the task into the extraction stage and the linking stage.

#### 4.2.1 Extraction Stage

Instead of finding possible entity links dynamically after the mentions are extracted, we rely on a static dictionary, containing mapping from various mention aliases to entities, to extract all possible links in advance using fuzzy string matching. This simplifies the labeling effort, while also reduces the computation time for both training and inference.

The performance would then heavily depend on the quality of the dictionary. We recursively trace Wikipedia's Redirection, which defines a mapping from a mention to an entity, and Disambiguation pages, which maps a mention onto multiple possible entities, to build the dictionary. Various rule-based clean-ups are also performed for the mentions, entities and the mapping.

#### 4.2.2 Linking Stage

The linking model then computes the similarity between the mention and its candidate entities. The mention tower is similar to the open world model, where we run the input document through a language model and pool the outputs to get embeddings for the mentions. On the entity side, its Wikipedia texts are summarized offline into embeddings. For each mention-entity pair, the mention embedding is broadcasted to dot with its candidate entity embeddings after a linear projection, to output a relevance score, as shown in Figure 1c.

We also experimented with first predicting a mention score as in the open world case, but found little difference in the final entity metric. Additional supervision on salience is also added for entities based on the number of votes received from the raters. We concatenate these scores with some counter-based features such as the prior of the mention-entity link, to get the final linking score after feed forward layer.

## 5 Scaling Challenges

To have a good coverage over various documents, our system needs to scale across languages, entity types and document types. Naively, we can develop a model for each triple (language, entity type, document type) and run a combination of models for each piece of document. However, this would bring significant overhead in model development and model serving. Therefore, our system tackles these scaling challenges with the following techniques and train a single model instead.

### 5.1 Cross Language Model and Fine-Tuning

Transformer (Vaswani et al., 2017) based pre-trained language model has led to strong improvements on various natural language processing tasks (Wang et al., 2018). With cross-lingual pretraining, XLM (Lample and Conneau, 2019) can achieve state-of-art results cross languages. In our work, we employ XLM and further improve the prediction by fine-tuning on multilingual data. We compare the performance of zero-shot and fine-tuned product extraction models on ads in Table 2. While the zero-shot model predicts reasonably for Romance languages, e.g. French (fr), Portuguese (pt), it has a poor performance for Arabic (ar) and Vietnamese (vi). This is expected since the latter have very different characteristics from English. By fine-tuning on all-language data, we see a substantial boost in model performance for all languages.

|  | Zero-Shot | | | Fine-Tuned | | |
|---|---|---|---|---|---|---|
| Language | Precision | Recall | F1 | Precision | Recall | F1 |
| ar | 0.2556 | 0.0676 | 0.1069 | 0.3170 | 0.5331 | 0.3976 |
| da | 0.2437 | 0.4037 | 0.3040 | 0.4093 | 0.5444 | 0.4673 |
| de | 0.2966 | 0.3670 | 0.3281 | 0.3349 | 0.5921 | 0.4279 |
| en | 0.4301 | 0.6750 | 0.5254 | 0.4251 | 0.7036 | 0.5300 |
| es | 0.2739 | 0.3500 | 0.3073 | 0.3439 | 0.5955 | 0.4360 |
| fr | 0.3499 | 0.3584 | 0.3541 | 0.4067 | 0.5988 | 0.4844 |
| it | 0.3157 | 0.3626 | 0.3375 | 0.4152 | 0.6146 | 0.4956 |
| nl | 0.2466 | 0.4673 | 0.3228 | 0.3316 | 0.5299 | 0.4079 |
| pt | 0.3075 | 0.4395 | 0.3618 | 0.4122 | 0.6555 | 0.5061 |
| ru | 0.3144 | 0.4467 | 0.3691 | 0.4300 | 0.7021 | 0.5334 |
| vi | 0.1886 | 0.0283 | 0.0492 | 0.3653 | 0.6888 | 0.4774 |
| Overall | 0.3315 | 0.3834 | 0.3556 | 0.3861 | 0.6331 | 0.4797 |

Table 2: Multilingual fine-tuning of product name extraction model. Zero-shot model is trained on English only; fine-tuned model is trained on all languages (one-tenth of English sample size for each new language).

## 5.2 Multi-Task Learning For Extraction, Clustering, and Linking

Multi-task learning (Caruana, 1997) is a subfield of machine learning, in which multiple tasks are simultaneously learned by a shared model. Such approaches offer advantages like improved data efficiency, reduced overfitting through shared representations, and fast learning by leveraging auxiliary information. It has been proved effective in various applications like Computer Vision (Zhang et al., 2014) and Natural Language Processing (Vaswani et al., 2017). In previous subsections, we train models separately and predict in parallel for different entity types. This is advantageous in that we can train a model for a new entity type or update the model for an existing entity type without affecting other entity models. However, this causes ever-increasing inference costs as new entity types are considered. Currently we have 5 entity types and 7 Transformer-based models, which means to run 7 XLM encoders for every ad, web page, etc. The heavy inference cost is a major blocker for our service. To resolve this issue, we developed the unified model structure and training framework. We are able to co-train all entity extraction and linking models with a shared XLM encoder. Since the encoding part accounts for the majority of all computation, the inference time is reduced to 1/7 of before and unblocks the service. Table 3 displays the performance of the shared-encoder models trained with the framework. It can be seen that they have a performance comparable with that of separately trained models. While the closed world linking model has a slightly better accuracy with co-training, the product name extraction model performs slightly worse. This is probably because a single XLM of a moderate size may not encode all info required by different entity extraction heads. We expect increasing the capacity of encoder will reduce the conflicts. To sum up, the unified model permits new entity types with little inference cost and only slight performance drop.

| Task | Metric | Separate Models | Shared-Encoder Models |
|---|---|---|---|
| Extraction | Precision | 0.4301 | 0.4171 |
|  | Recall | 0.6750 | 0.6671 |
|  | F1 | 0.5254 | 0.5133 |
| Closed World | Accuracy | 0.6729 | 0.6815 |

Table 3: Co-train product name extraction and closed world linking models with a shared XLM encoder

| Task | Metric | Ads Model | | Ads+Web Pages Model | |
|---|---|---|---|---|---|
|  |  | Ads | Web Pages | Ads | Web Pages |
| Extraction | Precision | 0.4301 | 0.4519 | 0.4315 | 0.5148 |
|  | Recall | 0.6750 | 0.5167 | 0.6951 | 0.6906 |
|  | F1 | 0.5254 | 0.4821 | 0.5325 | 0.5899 |
| Closed World | Accuracy | 0.6729 | 0.6106 | 0.6811 | 0.6852 |

Table 4: Transfer learning between product name extraction and closed world linking models between ads and web pages data. The first model is trained on ads only; the second is trained on both ads and web pages. The sample sizes of ads and web pages are the same.

## 5.3 Cross Document Transfer Learning

Transfer learning aims at improving the performance of target models on target domains by transferring the knowledge contained in different but related source domains (Zhuang et al., 2021). Different transfer learning approaches are developed from zero-shot transfer learning (Xian et al., 2017) to few-shot transfer learning (Vinyals et al., 2016). We incorporate the transfer learning framework in our system to solve cross document types challenge. We run experiments on zero-shot transfer learning and few-shot transfer learning as in Table 4. As we can see, the transfer learning could boost the performance of the model on both document types.

## 6 Conclusion And Future Work

In this paper, we present the platform of the entity extraction at giant internet company's scale. We discuss the practical learnings from our work. In the future, we would like to improve the efficiency of Transformer related language model as discussed in (Tay et al., 2020).